\documentclass[runningheads]{llncs}
\usepackage{graphicx}
\usepackage{amsmath}
\usepackage{amssymb}
\usepackage{booktabs}
\usepackage{multirow}
\usepackage{subfig}
\usepackage[misc,geometry]{ifsym}
\usepackage{lipsum}

\newcommand\blfootnote[1]{%
  \begingroup
  \renewcommand\thefootnote{}\footnote{#1}%
  \addtocounter{footnote}{-1}%
  \endgroup
}

\begin{document}
\title{Multi-dimensional Fusion and Consistency for Semi-supervised Medical Image Segmentation}
\titlerunning{ }
%

\author{Yixing Lu\inst{1} \and Zhaoxin Fan\inst{2}\thanks{Equal Contribution.} \and Min Xu\inst{2}\textsuperscript{\Letter}}
\authorrunning{Y. Lu et al.}
%
\institute{
  University of Liverpool, Liverpool, United Kingdom \and Mohamed bin Zayed University of Artificial Intelligence, Abu Dhabi, United Arab Emirates\\
\email{xumin100@gmail.com}
}

\maketitle              
\blfootnote{We thank Bowen Wei for helpful discussions on this work.}
\begin{abstract}
In this paper, we introduce a novel semi-supervised learning framework tailored for medical image segmentation. Central to our approach is the innovative Multi-scale Text-aware ViT-CNN Fusion scheme. This scheme adeptly combines the strengths of both ViTs and CNNs, capitalizing on the unique advantages of both architectures as well as the complementary information in vision-language modalities. Further enriching our framework, we propose the Multi-Axis Consistency framework for generating robust pseudo labels, thereby enhancing the semi-supervised learning process. Our extensive experiments on several widely-used datasets unequivocally demonstrate the efficacy of our approach.

\end{abstract}
\section{Introduction}

Medical image segmentation is a pivotal and intricate process within the realm of intelligent diagnosis, entailing the extraction of regions of interest within medical imagery. This task is of paramount importance for enabling precise diagnosis and tailored treatment. Over recent years, Convolutional Neural Networks (CNNs) \cite{huang2020unet,guo2021sa,cai2020dense} and Vision Transformers (ViTs) \cite{cao2021swin,medT,gao2021utnet}, both endowed with a U-shaped architecture, have witnessed significant advances in the domain of medical image segmentation.

Medical image segmentation literature mainly employs pretrained Convolutional Neural Networks (CNNs) \cite{huang2020unet} or Transformers \cite{cao2021swin}. The benefits of using both CNNs and Vision Transformers (ViTs) haven't been thoroughly explored. Interestingly, CNNs and ViTs seem to complement each other for medical image understanding. CNNs excel in local feature recognition \cite{baker2020local}, while ViTs are superior in comprehending long-range dependencies \cite{unetr,transunet}. For medical image segmentation, combining these strengths is crucial to understanding the organ and its interrelations with others. This raises the question: \emph{\textbf{Can we fuse the strengths of CNNs and ViTs into a single framework for medical image segmentation?}} An additional noteworthy observation is that both Convolutional Neural Networks (CNNs) and Vision Transformers (ViTs) tend to necessitate extensive quantities of annotated data for effective training. While this may be manageable in the realm of natural image segmentation, it poses a formidable hurdle in the context of medical image segmentation, where the process of annotation is both laborious and costly. As a result, a second question emerges: \emph{\textbf{Is it feasible to reduce the model's dependency on annotated medical image-mask pairs without undermining the performance of the ViT/CNN?}}

In response to the aforementioned questions, we introduce a novel semi-supervised learning framework for medical image segmentation in this study. We first propose a simple yet efficacious Multi-scale ViT-CNN Fusion scheme. The underlying principle of this approach is that the integration of ViT and CNN features can equip the model with the ability to capture both intricate local details and extensive global long-range dependency information. Additionally, given that both the CNN and ViT are pretrained on large-scale networks, they can retain both abstract natural features and domain-specific medical features during fusion, thereby further enhancing the segmentation task. Moreover, inspired by vision-language models \cite{denseclip,clipseg,cris,xu2021simple}, and considering the relative ease of obtaining text descriptions for medical images, we introduce a text-aware language enhancement mode to further enrich the learned features. This effectively addresses the first question. Subsequently, we incorporate a Multi-Axis Consistency framework in our study to extend our approach to scenarios where annotated labels are limited. Within this framework, we unify and formulate multiple consistency regularizations under a single framework, dubbed Multi-AXIs COnsistency (MaxiCo). This framework combines intra-model, inter-model, and temporal consistency regularizations to generate robust probability-aware pseudo-labels, thereby enabling the use of a large corpus of unlabeled data for semi-supervised network training. Furthermore, we design a voting mechanism within this module, whereby each intermediate output can contribute to the final pseudo-label. This mechanism further enhances the trustworthiness of pseudo-labels and bolsters the final model's performance, therefore providing a satisfactory answer to the second question.

To deliver a comprehensive demonstration of the efficacy of our proposed method, we have undertaken extensive experimentation using the  MoNuSeg \cite{monuseg} dataset and QaTa-COV19 \cite{degerli2022osegnet} dataset. The empirical results obtained from these experiments substantiate that our method establishes a new benchmark in fully-supervised settings, outperforming existing state-of-the-art methodologies. Moreover, within semi-supervised scenarios, our strategy shows remarkable superiority over other leading-edge techniques.  

Our contribution can be summarized as:  1) We pioneer a semi-supervised framework that harnesses the power of textual information to support fused ViT-CNN networks for medical image segmentation, representing a unique approach to this problem. 2) We propose a novel Multi-scale Text-aware ViT-CNN Fusion methodology that adroitly amalgamates CNNs and ViTs to boost segmentation accuracy.  3) We introduce a novel Multi-Axis Consistency Learning module that capitalizes on consistency regularizations to generate reliable pseudo-labels for semi-supervised learning, effectively addressing the issue of data scarcity.
\section{Related Work}
\noindent \textbf{Transformers in Medical Image Segmentation.}
The success of Vision Transformer (ViT) \cite{vit} in various computer vision tasks has led to its integration into medical image segmentation \cite{transseg,gao2021utnet,xie2021cotr,zhang2021transfuse,li2023lvit}. Some studies use transformers for image representation \cite{unetr}, while others propose hybrid encoders combining transformers and convolutional neural networks (CNNs) \cite{transunet}. Cao et al. \cite{cao2021swin} proposed a pure transformer network replacing convolutional layers with Swin Transformer blocks \cite{swin} and the traditional Unet skip-connection with a transformer-based channel-attention module \cite{uctransnet}. However, these methods are typically trained in a fully-supervised manner, which may be impractical due to the scarcity of annotated medical data. To address this, we introduce a semi-supervised approach that fuses ViT-CNN networks for medical image segmentation, aiming to overcome the challenge of limited annotated data.

\noindent \textbf{Semi-Supervised Medical Image Segmentation.}
In light of the challenge posed by the dearth of annotated data in medical image segmentation, semi-supervised learning has come to the fore as a promising solution \cite{semi1,semi2,semi3,semi4}. Predominant strategies for semi-supervised learning encompass pseudo labeling \cite{pseudo}, deep co-training \cite{cotrain}, and entropy minimization \cite{entropy}. In our work, we adopt a consistency learning framework to generate pseudo labels for unmarked images.  There are several versions of consistency regularization methods, including temporal consistency \cite{temporal}, model-level consistency \cite{crossteaching}, and pyramid consistency \cite{pyramid}, in existing literature. However, most of these methods only depend on a single type of regularization, which will limit the porwer of the model. In contrast, our approach amalgamates multiple consistency regularizations and employs a voting mechanism to produce more robust pseudo labels, which demonstrates better performance.

\noindent \textbf{Vision-Language Fusion for Dense Predictions.}In recent years, the fusion of vision and language in large-scale pretraining has garnered significant attention. The CLIP model \cite{radford2021learning} showcases this, demonstrating impressive transfer learning across multiple datasets. Building on this, researchers have explored fine-tuning the CLIP model for dense prediction tasks \cite{denseclip,li2022language,cris}, framing it as a pixel-text matching problem. Vision-language models also enable zero-shot inference, bridging the gap between seen and unseen categories \cite{xu2021simple}. Some research has further explored the potential of visual prompts and their interpolation with text prompts during training \cite{clipseg}. In this paper, we use textual information to enhance ViT-CNN training for medical image segmentation, showcasing a novel application of vision-language fine-tuning.

\section{Methodology}

\subsection{Overview}

Given an image $I \in X$, where $X$ represents the space of all possible medical images, and a text input $T \in T$, where $T$ is the space of all possible text inputs (e.g., medical notes or labels), the task of medical image segmentation in our study is to learn a mapping function $F_\theta: X \times T \rightarrow Y$:

\begin{equation}
F_\theta: X \times T \rightarrow Y  
\end{equation}

Where $Y$ represents the segmentation masks corresponding to the input medical image, and $\theta$ denotes the parameters of our model. The goal is to train the model such that the mapping function $F_\theta$ can accurately predict the segmentation mask $y \in Y$ for any given input image and text $(I, T)$.

To address the first challenge, our Multi-scale Text-aware ViT-CNN Fusion scheme integrates a pretrained ViT and CNN, incorporating text features for increased prediction accuracy. We perform vision-language pretraining to obtain vision and text features, aligning them to formulate ViT features. These are then fused with CNN features at various resolutions, enabling the efficient use of local and global features.

To facilitate semi-supervised training, we introduce a Multi-Axis Consistency framework to generate pseudo labels, leveraging inter-model, multi-scale intra-model, and temporal consistency. Our network makes multiple predictions in a single pass, generating probabilistic pseudo labels via a voting mechanism, supporting semi-supervised training.
\begin{figure}[h]
    \centering
    \includegraphics[width=\textwidth]{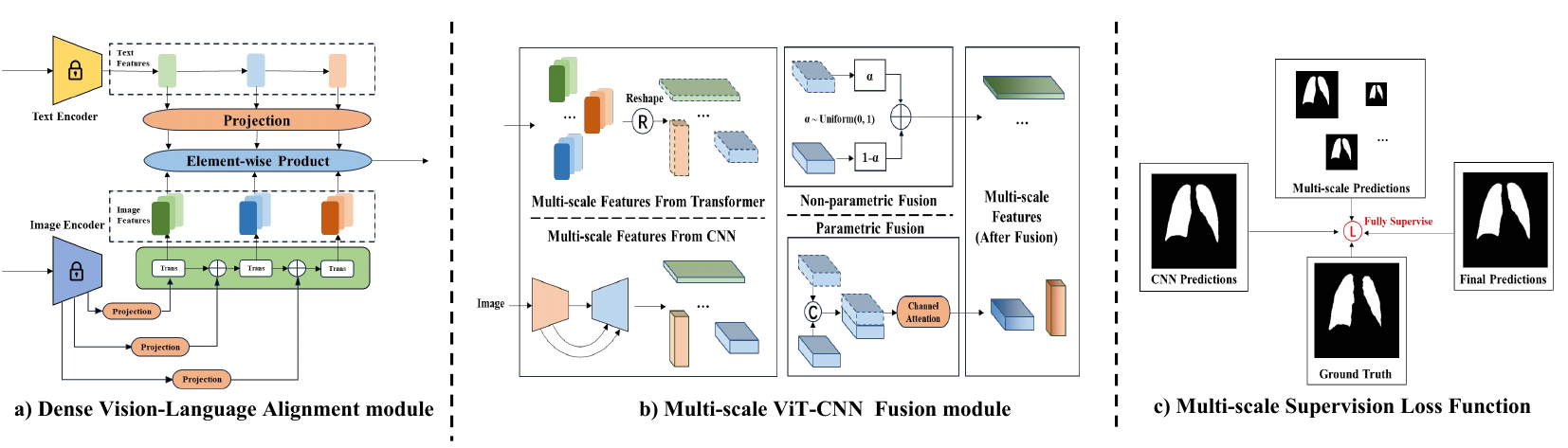}
    \vspace{-0.25in}
    \caption{The illustration of Multi-scale Text-aware ViT-CNN Fusion.}
    \label{fig:module1}
    \vspace{-0.5in}
\end{figure}

\subsection{Multi-scale Text-aware ViT-CNN Fusion}
In this section, we present a novel architectural design named Multi-scale Text-aware ViT-CNN Fusion, as depicted in Fig. \ref{fig:module1}. This scheme is primarily composed of three major components: \emph{Dense Vision-Language Alignment module}, \emph{Multi-scale ViT-CNN Fusion module}, and a \emph{Supervised Loss function} for joint training. The Dense Vision-Language Alignment model is responsible for aligning the vision and text features into a common embedding space. By performing this alignment, we can effectively exploit the complementary information from both modalities to enhance the feature representations. The second component, Multi-scale ViT-CNN Fusion module, facilitates the fusion of the features extracted by the ViT and the CNN. This fusion is carried out at multiple scales, allowing the model to capture abstract features, domain-specific features, local details, and global long-range dependencies at different resolutions. Finally, the Multi-scale Supervision Loss Function. By optimizing this loss, the network learns to predict segmentation masks in a multi-scale and progressive manner. Next, we introduce them in detail.

\noindent \textbf{Dense Vision-Language Alignment module.}  In our work, we incorporate both image and text information as inputs for segmentation. The inclusion of text allows us to capture the strengths of transformers more effectively and bolsters the fusion of Vision Transformers (ViTs) and Convolutional Neural Networks (CNNs). This approach leverages contextual cues from text to enhance segmentation precision. To align the image and text features, we adopt a progressive approach. Visual features are extracted from a sequence of layers (${3, 6, 9, 12}$) of a pretrained visual encoder, forming a set $L={x_{1},x_{2},...,x_{\ell-1},x_{\ell}}$ where $x_1$ and $x_{\ell}$ represent the features from the shallowest and deepest layers, respectively. We obtain text embeddings, denoted by $y$, from a pretrained clinical text encoder.

We use transformer layers with skip connections to compactly represent visual features, as shown in the equation below:

\begin{equation}
\begin{split}
X_i = \begin{cases}
\mathsf{TransLayer}_{i}(W^x_i x_i), &i=\ell \\
\mathsf{TransLayer}_i(W^x_i x_{i}+X_{i+1}), &i\in L-\{\ell\}
\end{cases}
\end{split}
\end{equation}

Here, $W^x_i$ is a linear layer for dimension reduction, and $\mathsf{TransLayer}_i$ denotes a Transformer layer. We reduce the dimension of the text embeddings and transfer them to different layers using simple MLP blocks:

\begin{equation}
\begin{split}
Y_i = \begin{cases}
    W^y_i y, &i=\ell \\
    W^y_i Y_{i+1}, &i\in L-\{\ell\}
\end{cases}
\end{split}
\end{equation}

Here, $W^y_i$ signifies a linear layer for dimension reduction, while $W_i^y$ denotes an MLP block for transferring text embeddings. We perform Vision-Text alignment using element-wise multiplication so that the alignment is "dense":

\begin{equation}
        Z_i=W^X_i X_i \odot W^Y_i Y_i, \quad i\in L
\end{equation}

In the equation above, $\odot$ represents element-wise multiplication, and $W^X_i$ and $W^Y_i$ are tensor reshape operations for $X_i$ and $Y_i$, respectively.

\noindent \textbf{Multi-scale ViT-CNN  Fusion module.} ViTs and CNNs each have their unique strengths in image analysis tasks. ViTs excel in capturing global dependencies, while CNNs are particularly adept at extracting local features. However, when dealing with complex tasks such as medical image segmentation, a combination of these two can be beneficial, leveraging global contextual understanding and local feature extraction.  Addressing this, we propose a dense fusion of ViT and CNN features at different resolutions. This approach is designed to enhance local interactions and preserve global knowledge. Our fusion method follows two guiding principles: 1) it should improve model performance, and 2) it should maintain the robustness of each individual feature to avoid over-dependence on either. 

We begin with a non-parametric fusion method, where the fusion parameter $\beta$ is uniformly sampled from $[0, 1]$. A Unet CNN processes the input $I\in \mathbb{R}^{H\times W\times 3}$, projecting it initially to $C_1$ and then applying $(N-1)$ down/up-sampling operations to yield multi-scale features $F_j^{CNN}\in \mathbb{R}^{H_j\times W_j\times C_j}$ at $N$ different resolutions ($N=4$ in our case). 

ViT features $Z_i$ are projected to match the size of the corresponding CNN features $F_j^{CNN}$, resulting in the ViT features $F^{ViT}_j$. These are then fused with the CNN features as follows:

\begin{equation}
        F_j = \beta F^{CNN}_j + (1-\beta) F^{ViT}_j, \quad j=1,2,...,N
\end{equation}

$\beta$ is sampled from $[r_1,r_2] (0\leq r_1 <r_2 \leq 1)$, and $F_j$ is the fused feature map at level $j$.

Beyond non-parametric fusion, we explore parametric fusion, employing a channel attention mechanism \cite{xcit} at each scale. This mechanism is defined as:

\begin{equation}
    \centering
    \begin{split}
    &\hat{X}=W\cdot\mathsf{Attention}(\hat{Q}, \hat{K}, \hat{V})\\
            &\mathsf{Attention}(\hat{Q}, \hat{K}, \hat{V})=\hat{V}\cdot\mathsf{Softmax}(\frac{\hat{K}\cdot\hat{Q}}{\alpha})
    \end{split}
\end{equation}

Here, $\hat{X}\in \mathbb{R}^{H\times W\times C}$ denotes the output feature map; $\hat{Q} \in \mathbb{R}^{HW\times C}, \hat{K}\in \mathbb{R}^{C\times HW}, \hat{V} \in \mathbb{R}^{HW\times C}$ are tensors reshaped from $Q, K, V$, respectively; $W$ is a $1\times 1$ convolution for output projection; $\alpha$ is a learnable parameter to control the magnitude of $\hat{K}\cdot\hat{Q}$. The definition of $Q, K, V$ is $Q, K, V = W^QX, W^KX, W^VX$ and $X = \mathsf{LayerNorm}([F^{CNN}, F^{VIT}])$, where $[\cdot,\cdot]$ denotes feature concatenation and $W^{(\cdot)}$ denotes point-wise $1\times1$ convolutions.


\noindent \textbf{Multi-scale Supervision Loss Function.}
In our proposed network architecture that combines ViT and CNN in a multi-scale manner, multiple predictions are generated in a single forward pass. However, relying exclusively on the final output for training can lead to convergence issues. To address this, we propose an end-to-end optimization of multiple predictions, which we term the Multi-scale Supervision Loss function.

We denote the network's final prediction as $P$ and its multi-scale predictions as $S={Q_1,Q_2,...,Q_{s-1},Q_s}$, where $Q_s$ represents the prediction at the $s$-th scale. For simplicity, we exclude the final prediction $P$ from the set $S$. The CNN branch prediction is represented by $R$. Then, we utilize multi-scale predictions $S+P$ and the CNN output $R$ to compute the loss. The Multi-scale Supervision Loss function is formulated as follow:
\begin{equation}
\mathcal{L}_{ms} = \alpha _1 \mathcal{L}(P, T) + \alpha _2 \mathcal{L}(R, T)+ \alpha _3 \frac{1}{|S|}\sum _{s=1}^{|S|} \mathcal{L}(Q_s,T)
\end{equation}

Here, $\mathcal{L}$ refers to the average of Dice loss and Cross Entropy loss. $T$ denotes the ground truth label, and $|S|$ is the cardinality of set $S$. The weights $\alpha_1, \alpha_2,$ and $\alpha_3$ are used to balance each term in the loss function, and are set to $\alpha_1=\alpha_2=1$ and $\alpha_3=0.6$ for all our experiments.

\begin{figure}[h]
    \centering
    \includegraphics[width=\textwidth]{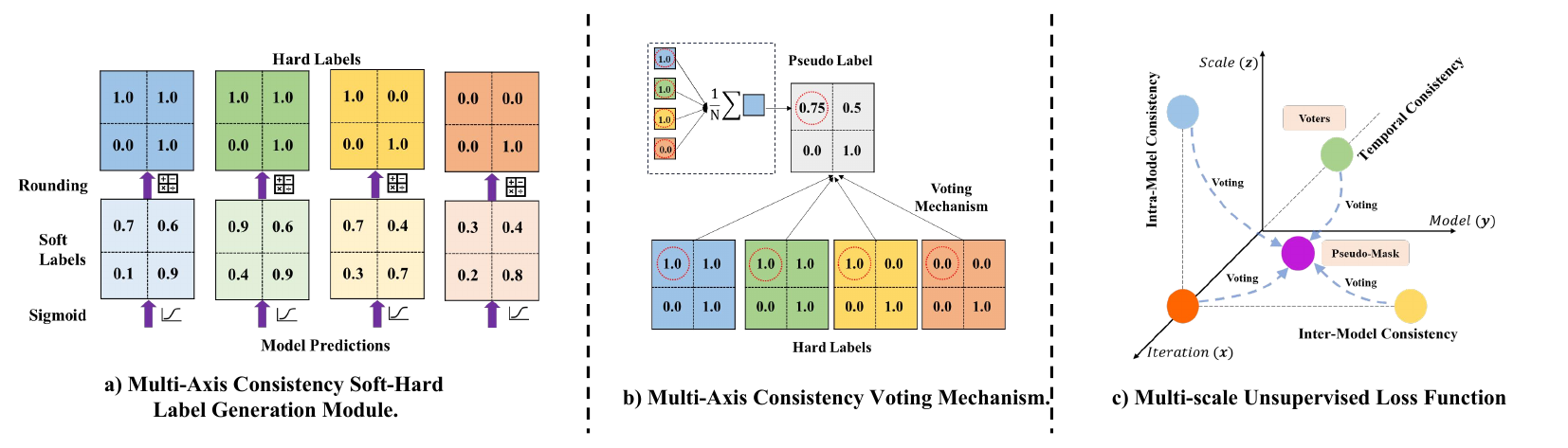}
     \vspace{-0.25in}
    \caption{The illustration of Multi-Axis Consistency Framework.}
    \label{fig:module2}
    \vspace{-0.25in}
\end{figure}
\vspace{-0.2in}
\subsection{Multi-Axis Consistency Framework}
In our pursuit to accomplish semi-supervised learning, we present a novel Multi-Axis Consistency framework as illustrated in Fig. \ref{fig:module2}. This all-encompassing framework is made up of three main components: the \emph{Multi-Axis Consistency Soft-Hard Label Generation Module, the Multi-Axis Consistency Voting Mechanism, and the Multi-scale  Unsupervised Loss Function}. The Soft-Hard Label Generation Module generates robust labels, taking into consideration intra-model and inter-model consistency, as well as temporal consistency. The Consistency Voting Mechanism selects the most probable predictions across different models and scales, thereby enhancing the robustness and accuracy of the learning process. The Multi-scale Unsupervised Loss Function provides a metric for model optimization in scenarios where ground truth labels are absent, promoting the extraction of valuable features from unlabeled data. Next, we introduce them in detail.

\noindent \textbf{Multi-Axis Consistency Soft-Hard Label Generation Module.}  This innovative module is designed based on the Coordinate Systems we established to model different consistency paradigms, as depicted in Fig. \ref{fig:module2} (a). We represent the input as $x$ and the consistency condition as $\theta=[m, s, t]^T$, indicating that the output is generated by model $m$, scale $s$, and training iteration $t$. The module's objective is to minimize the distance between two outputs under consistency regularization from multiple axes. The module achieves this by applying an augmentation $\sigma$ to the input $x$ and generating a modified input $\hat{x}$ while ensuring a small consistency relaxation $\epsilon$. This process is expressed as follows:

\begin{equation}
\begin{split}
    &\min \lVert f(x,\theta) - f(\hat{x}, \theta +\epsilon)\rVert \\
&s.t.  \hat{x}=\sigma (x), \lVert \epsilon \rVert \rightarrow 0 \\
\end{split}
\end{equation}

Then, the module generates robust labels by predicting multiple segmentation maps $P_\theta \in \mathbb{R}^{H\times W\times K}, \theta \in \Theta$, where $K$ denotes the number of segmentation classes. These are the soft labels. The module then converts these soft labels into binary hard labels using a threshold. Next, we introduce the whole process in our ,multi-axis consistency voting mechanism.

\noindent \textbf{Multi-Axis Consistency Voting Mechanism.} The Voting Mechanism is implemented based on the Semi-Supervised Learning strategy, illustrated in Fig. \ref{fig:module2} (b). This mechanism samples a subset of the predicted segmentation maps within the consistency relaxation and utilizes them to generate a probabilistic pseudo-label. It leverages the outputs from the Vision Transformer (ViT), Convolutional Neural Network (CNN), and multi-scale outputs to collaboratively vote for the most probable pseudo-label. The pseudo-label includes the probability of each pixel belonging to a specific class.

To achieve this, the mechanism first predicts multiple segmentation maps, which can be considered as "soft labels" indicating the probability of each class. These soft labels are then converted into binary "hard labels" using a threshold of 0.5, as shown in the first part of the equation.

\begin{equation}
    \begin{split}
         M_\theta(h,w,k) &=  \left\{ \begin{array}{rcl}
             1 & \mbox{for} & P_{\theta}(h,w,k)\geq 0.5 \\
             0 & \mbox{for} & P_{\theta}(h,w,k)<0.5
        \end{array}\right.
    \end{split}
\end{equation}
Here, $M_\theta(h,w,k)$ denotes the binary hard label of pixel $(h,w)$ for class $k$ under condition $\theta$ and $P_{\theta}(h,w,k)$ represents the soft label corresponding to the same. The final pseudo-label $M_{pseu}$ is then generated by taking the average of these binary hard labels across all conditions $\theta \in \Theta$, as expressed in the second part of the equation.

\begin{equation}
    \begin{split}
        M_{pseu} &= \frac{1}{|\Theta|} \sum_{\theta \in \Theta}{M_\theta}
    \end{split}
\end{equation}
In this equation, $M_{pseu}$ represents the final probabilistic pseudo-label and $|\Theta|$ denotes the cardinality of set $\Theta$. In this way, the Multi-Axis Consistency Voting Mechanism generates robust probabilistic pseudo-labels that reflect the consensus among different models (ViT, CNN) and multi-scale outputs, embodying the concept of multi-axis consistency.

\noindent \textbf{Multi-scale Unsupervised Loss Function.} The unsupervised loss function is incorporated in our semi-supervised training via an unsupervised loss term, as is shown in Fig. \ref{fig:module2} (c). For unlabeled images, the function first generates pseudo-labels according to multiple network outputs. Multiple outputs from the current training iteration as well as previous iterations all contribute to the generation of the pseudo-label. The function aims to minimize the distance between contributors from the current training iteration to the pseudo-label:

\begin{equation}
    \mathcal{L}_{unsup} = \frac{1}{|\Theta|}\sum_{\theta \in \Theta} \mathcal{L}(P_\theta, M_{pseu})
\end{equation}

Here, $\mathcal{L}$ represents the average of Dice loss and Cross-Entropy loss. The final loss for semi-supervised learning $\mathcal{L}_{final}$ is represented by the weighted sum of $L_{sup}$ and $L_{unsup}$, as shown below:

\begin{equation}
    \mathcal{L}_{final} = \mathcal{L}_{sup}+\lambda \mathcal{L}_{unsup}
\end{equation}

Here, $\lambda$ is a weight factor, defined by a time-dependent Gaussian warming-up function to control the balance between the supervised loss and unsupervised loss.

\section{Experiments}
\subsection{Experiment Setup}

We use pretrained vision and language transformers, which remain frozen during training. We use a ViT pretrained on the ROCO dataset \cite{roco} via DINO \cite{dino} as our vision backbone, and Clinical BERT \cite{clinicalbert} as our language backbone.  We adopt U-Net \cite{unet} as the CNN branch. We set the batch size to 4 and the initial learning rate to $10^{-3}$, using the Adam optimizer \cite{adam} with cosine annealing cyclic schedule. Data augmentation includes random flips and $90^{\circ}$ rotations. All experiments are conducted on an A5000 GPU. We use Dice and mIoU metrics as our evaluation metrics. The experiments are conducted on MoNuSeg \cite{monuseg} and QaTa-COV19 \cite{degerli2022osegnet} datasets. The MoNuSeg  dataset includes images of tissue from various patients with tumors and approximately 22,000 nuclear boundary annotations across 30 training images and 14 test images. And the QaTa-COV19 dataset includes 9258 annotated COVID-19 chest radiographs. The text annotations for both datasets are derived from \cite{li2023lvit}.

\begin{figure*}[t]
    \centering
    \begin{minipage}{.45\textwidth}
        \centering
        \scalebox{0.8}{ 
            \begin{tabular}{c c c c c}
                \hline
                \multirow{2}{*}{Method} & \multicolumn{2}{c}{MoNuSeg} & \multicolumn{2}{c}{QaTa-COV19} \\ \cline{2-5} 
                & Dice (\%) & mIoU (\%)  & Dice (\%) & mIoU (\%) \\ \hline
                Unet & 76.45 & 62.86 & 79.02 & 69.46 \\
                Unet++ & 77.01 & 63.04 & 79.62 & 70.25 \\
                AttUnet & 76.67 & 63.74 & 79.31 & 70.04 \\
                nnUnet & 80.06 & 66.87 & 80.42 & 70.81 \\
                MedT & 77.46 & 63.37 & 77.47 & 67.51 \\
                TransUnet & 78.53 & 65.05 & 78.63 & 69.13 \\
                GTUnet & 79.26 & 65.94 & 79.17 & 69.65 \\
                Swin-Unet & 77.69 & 63.77 & 78.07 & 68.34 \\
                UCTransNet & 79.87 & 66.68 & 79.15 & 69.60 \\ 
                \hline
                Ours+PF & 79.91 & 66.74 & \textbf{82.29} & \textbf{72.87} \\
                Ours+NPF & \textbf{80.60} & \textbf{67.66} & 82.03 & 72.80 \\ \hline
            \end{tabular}
        } 
        \captionof{table}{Results under Fully-supervised Learning and Comparison with the state-of-the-arts. ``PF" and ``NPF" represent Parametric fusion and Non-Parametric Fusion, respectively. Different values of $\beta$ during inference are also included.}
        \label{tab:tb1}
    \end{minipage}%
    \hspace{1cm} 
    \begin{minipage}{.45\textwidth}
        \centering
        \begin{tabular}{cccc}
            \hline
            \multirow{2}{*}{Setting} & \multirow{2}{*}{Labels (\%)} & \multicolumn{2}{c}{MoNuSeg} \\ \cline{3-4} 
            & & Dice (\%) & mIoU (\%) \\ \hline
            \multirow{3}{*}{PF} & 25 & \underline{78.59} & \underline{64.99} \\
            & 50 & 78.85 & 65.36 \\
            & 100 & 79.91 & 66.74 \\ \hline
            \multirow{3}{*}{NPF} & 25 & 78.47 & 64.88 \\
            & 50 & \underline{79.26} & \underline{65.94} \\
            & 100 & \underline{80.16} & \underline{67.06} \\ \hline
        \end{tabular}
        \captionof{table}{Results under Semi-Supervised Setting. ``PF" and ``NPF" represent Parametric fusion and Non-Parametric Fusion, respectively.}
        \label{tab:tb2}
    \end{minipage}
     \vspace{-0.4in}
\end{figure*}

\subsection{Quantitative Results}
\label{subsec:results}
In this section, we conduct main experiments on MoNuSeg dataset on both fully-supervised and semi-supervised settings. We also include the fully-supervised results on QaTa-COV19 dataset.

\noindent \textbf{Results on Fully-Supervised Setting.} 
In our research, we compare our methodology to state-of-the-art methods in a fully-supervised setting, these methods include Unet\cite{unet}, Unet++\cite{unet++}, AttUnet\cite{attunet}, nnUnet\cite{nnunet}, MedT\cite{medT}, transUnet\cite{transunet}, GTUnet\cite{gtunet}, Swin-Unet\cite{cao2021swin}, and UCTransNet\cite{uctransnet}. The results of our approach and the other state-of-the-art methods in a fully-supervised learning setting are presented in Table \ref{tab:tb1}. Notably, our method demonstrates a significant improvement over existing approaches. With the employment of parametric ViT-CNN fusion, our method achieves results that are not only comparable with TransUnet \cite{transunet} on the MoNuSeg dataset but also surpasses it under specific conditions. More notably, our approach exhibits superior performance under non-parametric feature fusion, namely, random fusion with a uniformly sampled $\beta$ during training. In this case, our method sets a new benchmark on the MoNuSeg dataset, outperforming all other state-of-the-art methods. This remarkable performance demonstrates the robustness of the random fusion strategy, where both the ViT and CNN branches can learn strong representations.

\noindent \textbf{Results on Semi-Supervised Setting.}   In Table \ref{tab:tb2}, we present our results under a Semi-Supervised setting. These results are achieved by using 25\% and 50\% labels for model evaluation, under our proposed Multi-Axis Consistency framework. The performance of our method stands out in several ways: 1) Most notably, our method delivers a comparable result to TransUnet in a fully-supervised setting even with only 25\% labels. Moreover, when we possess 50\% labels, the result improves significantly. This clearly showcases the potential of our proposed method. It illustrates how our method can effectively reduce the reliance on labeled data by learning from limited data and large-scale unlabeled data, thereby alleviating the cost of labels. 2) Our method with ViT-CNN random fusion and parametric channel attention consistently produces strong results across all semi-supervised settings. While the version with ViT-CNN random fusion outperforms the version with parametric channel attention by a small margin (less than 0.5\%) when using 50\% and 100\% labels, the results are fairly comparable. This highlights the advantages of our multi-scale ViT-CNN fusion, underscoring its ability to capture both global and local interactions and retain pre-trained knowledge. All these findings reinforce the effectiveness and efficiency of our method in Semi-Supervised settings, demonstrating that it is a promising approach for future research and applications.

\subsection{Qualitative Comparisons}
\begin{figure*}[t]
  \centering
    \includegraphics[width=\textwidth]{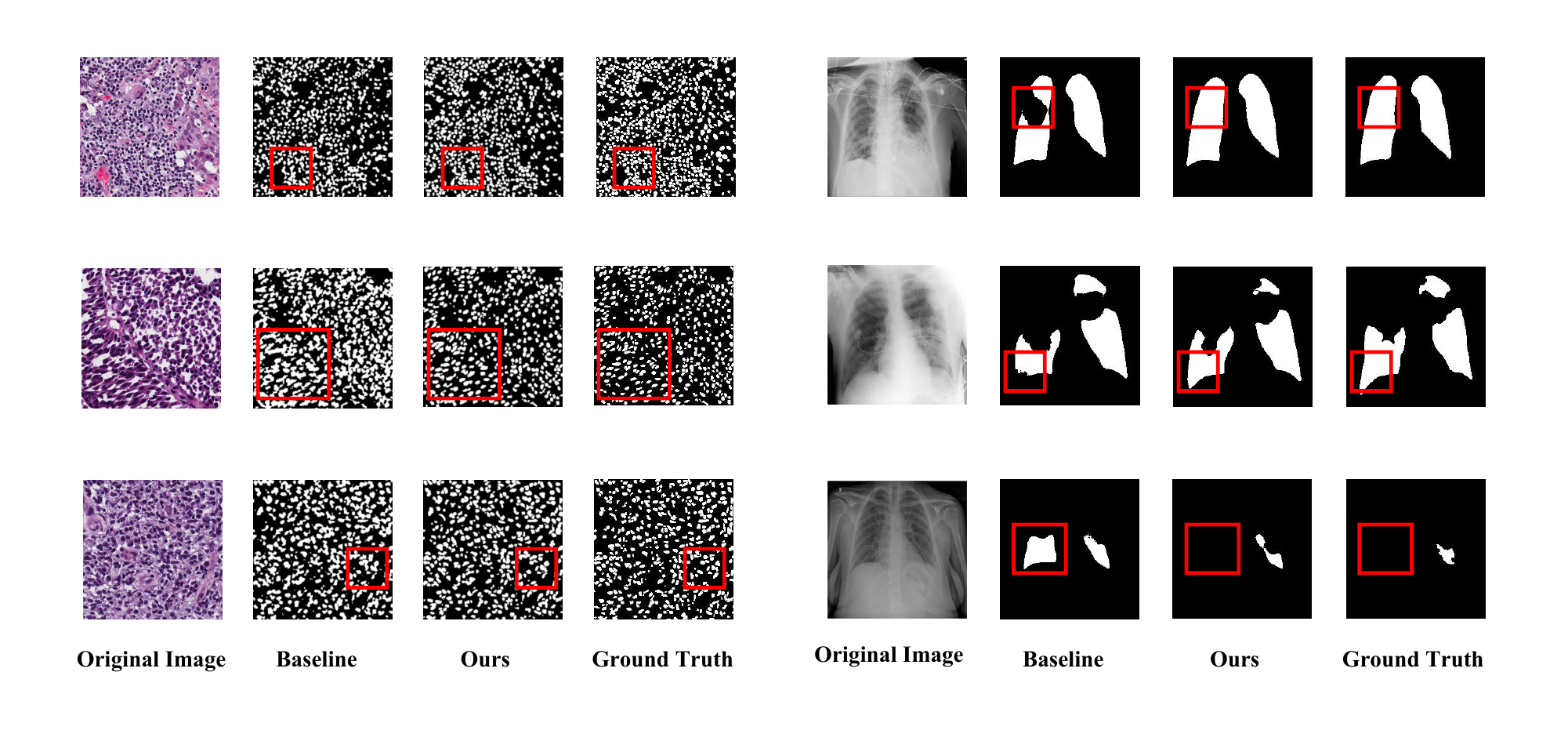}
     \vspace{-0.2in}
   \caption{Qualitative comparison. Left: Visualization results on MoNuSeg dataset. Right: Visualization results on QaTa-COV19 dataset.}
   \label{fig:fig7}
    \vspace{-0.3in}
\end{figure*}

Fig. \ref{fig:fig7} provides a visual comparison between our method and our baseline method \cite{clipseg}. Both sets of results are obtained under a fully-supervised training setting. To facilitate a clearer distinction between the two, we have highlighted specific areas in each sample within a red box. Upon close examination, it's evident that our proposed method offers superior results with respect to the precision of boundary delineation and the accuracy of shape representation. These improvements are most apparent within the highlighted regions, where our method's predictions exhibit finer detail and higher fidelity to the original structures. A key factor contributing to this enhanced performance is the introduction of our multi-scale text-aware ViT-CNN fusion. This innovative approach significantly improves local feature extraction within the target medical domain, allowing for more accurate and detailed segmentation results. This clearly demonstrates the advantage of our method over traditional approaches, and its potential for providing superior outcomes in complex medical image analysis tasks.

\subsection{Ablation Study}
\vspace{-0.3in}
\begin{table*}[htbp]
  \centering
  \subfloat[Ablation study on proposed modules under fully-supervised learning.]{
  \setlength\tabcolsep{2pt}
\resizebox{0.45\textwidth}{12mm}{
  \begin{tabular}{ccccccc}
\hline
\multirow{2}{*}{Method} & \multicolumn{4}{c}{ViT-CNN Dense Fusion}                   & \multicolumn{2}{c}{MoNuSeg} \\ \cline{2-7} 
                         & Multi-scale Arch. & Text & ViT-CNN Fusion & Loss & Dice (\%)    & mIoU (\%)    \\ \hline
Baseline                 &                  &      &                &                & 68.03             &   52.05           \\ \hline

& \checkmark                 &      &                &                & 76.30             & 61.92             \\

Ours & \checkmark   & \checkmark     &                &                & 78.47             & 64.67              \\

& \checkmark     & \checkmark     & \checkmark               &      & 79.14             & 65.78             \\

& \checkmark    & \checkmark     & \checkmark               & \checkmark  & \textbf{80.16}    & \textbf{67.06}             \\
\hline

\end{tabular}
}
\label{tab:tb3}
  }
  \qquad
  \subfloat[Ablation study on Multi-Axis Consistency framework under semi-supervised learning with 50\% training labels.]{
\resizebox{0.45\textwidth}{14mm}{
 \begin{tabular}{ccccccc}
\hline
\multirow{2}{*}{Setting} & \multicolumn{3}{c}{Multi-Axis Consistency}   & \multicolumn{2}{c}{MoNuSeg} \\ \cline{2-6} 
 & Intra-Model & Inter-Model & Temporal & Dice (\%)    & mIoU (\%)    \\ \hline
 Sup. Only & & & & 77.94 & 64.12 \\
 \hline
\multirow{5}{*}{Ours}
                    & \checkmark         &             &         & 78.68             & 65.11             \\

                 &             & \checkmark            &         & 78.52	& 64.92   \\

                   &            &             & \checkmark         & 78.01   & 64.15       \\

                    & \checkmark         & \checkmark    &          &  78.88	& 65.37       \\

       & \checkmark      &              & \checkmark          & 78.84 & 65.31           \\
         &       & \checkmark      & \checkmark          & 78.57	& 64.97       \\
              &   \checkmark    & \checkmark      & \checkmark          & \textbf{79.26}	& \textbf{65.94}      \\
\hline
\end{tabular}
  }
    \label{tab:tb4}
  }

  \caption{Ablation studies on proposed modules.}
   \vspace{-0.4in}
  \end{table*}
  
\noindent \textbf{Ablation Studies on Mutli-scale Text-aware ViT-CNN Fusion.} To evaluate our ViT-CNN Fusion, we employed the state-of-the-art vision-language transformer dense finetuning method as our baseline, which didn't perform well in medical image segmentation due to over-reliance on the pretrained backbone and lack of multi-scale dense features. To counter these issues, we propose a multi-scale text-aware ViT-CNN fusion for optimized pretrained transformers. An ablation study was conducted to ascertain the contribution of each component. This analysis involved sequentially introducing multi-scale architecture, dense vision-text alignment, ViT-CNN fusion, and a joint training loss function.  Table \ref{tab:tb3} shows the results, with significant Dice gains for each module: 8.27\% for the multi-scale architecture, 2.17\% for the vision-text alignment, 0.67\% for the ViT-CNN fusion, and 1.02\% for the joint training loss. The data underscores the effectiveness of each part of our method, particularly the substantial role of ViT-CNN Fusion in improving medical image segmentation tasks.

\noindent \textbf{Ablation Studies on Multi-Axis Consistency.} 
Our research introduces Multi-Axis Consistency, an innovative framework for generating robust pseudo labels in semi-supervised learning, by integrating different consistency regularization types. Table \ref{tab:tb4} displays our results: each consistency regularization type improves semi-supervised performance compared to a supervised-only setting, highlighting their importance in semi-supervised learning. Notably, peak performance is achieved when all three types are combined, demonstrating the effectiveness of the Multi-Axis Consistency framework. This comprehensive approach leads to superior performance in semi-supervised learning, marking a significant advancement in generating pseudo labels and improving model performance.

\begin{figure*}[ht]
 \vspace{-0.2in}
  \centering
  \begin{minipage}[b]{0.32\linewidth}
    \includegraphics[width=\linewidth]{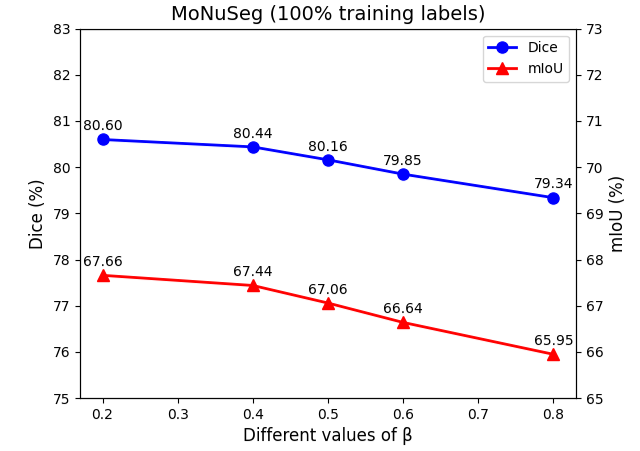}
    \caption{Impact of  different $\beta$ (full). }
    \label{fig:fig5_a}
  \end{minipage}
  \hfill
  \begin{minipage}[b]{0.32\linewidth}
    \includegraphics[width=\linewidth]{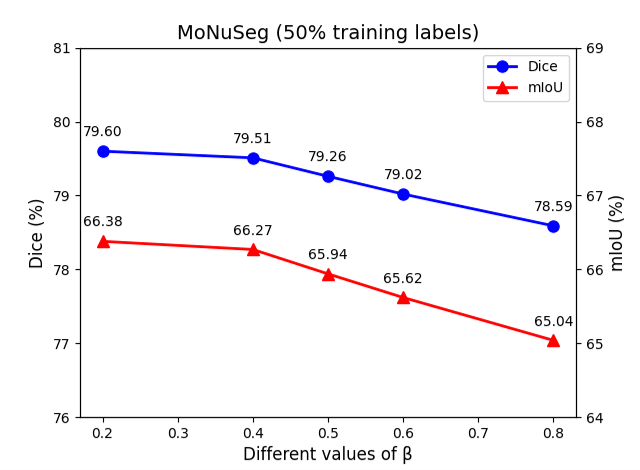}
    \caption{Impact of  different $\beta$(semi).}
    \label{fig:fig5_b}
  \end{minipage}
  \hfill
  \begin{minipage}[b]{0.32\linewidth}
    \includegraphics[width=\linewidth]{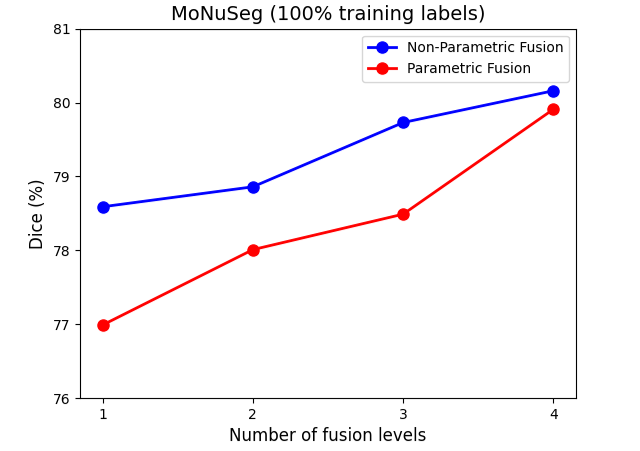}
    \caption{Impact of fusion level numbers.}
    \label{fig:fig5_c}
  \end{minipage}
  \vspace{-0.5in}
\end{figure*}

\noindent \textbf{In-depth Discussion on Multi-scale ViT-CNN Fusion.} In this section, we address two key questions experimentally: 1Why does ViT-CNN fusion work in semi-supervised settings? Our results (Table \ref{tab:tb3} and Fig. \ref{fig:fig5_a}) demonstrate this module's effectiveness in fully-supervised learning. Fig. \ref{fig:fig5_b} further shows that ViT-CNN fusion is crucial in semi-supervised settings, with performance increasing as $\beta$ decreases from 0.8 to 0.2. This suggests that both Transformer and CNN branches can independently perform well in such settings. 2) Why is multi-scale fusion important? We conducted ablation studies on fusion levels using different approaches: Non-Parametric Random Fusion and Parametric Channel attention. Fig. \ref{fig:fig5_c} shows that increased feature fusion levels improve model performance, underscoring the importance of multi-scale dense features in medical image segmentation and the effectiveness of our proposed multi-scale ViT-CNN fusion method.

\vspace{-0.1in}
\section{Conclusion}
In this paper, we propose a novel semi-supervised learning framework for medical image segmentation. In our work, a Text-aware ViT-CNN Fusion scheme is proposed to take advantages of both pretrained ViTs and CNNs as well as extracting both abstract features and medical domain specific features. Besides, a novel Multi-Axis Consistency framework is proposed to vote for pseudo label to encourage semi-supervised training. Experiments on serveral widely used datasets have demonstrated the effectiveness of our method.

%
%
%

\bibliographystyle{splncs04}
\bibliography{mybibliography}

\end{document}